\documentclass[preprint, 12pt]{elsarticle}

\usepackage{amssymb}
\usepackage{amsmath}
\usepackage{floatrow}
\usepackage{subfig}
\usepackage{graphicx}
\usepackage{cleveref}
\usepackage{makecell}
\usepackage{url}

\journal{Biocybernetics and Biomedical Engineering}

\begin{document}

\begin{frontmatter}



\title{Cellpose+, a morphological analysis tool for feature extraction of stained cell images}

\author[1]{Israel A. Huaman}

\author[1]{Fares D.E. Ghorabe}

\author[3]{Sofya S. Chumakova}

\author[1]{Alexandra A. Pisarenko}

\author[2]{Alexey E. Dudaev}

\author[2]{Tatiana G. Volova}

\author[2]{Galina A. Ryltseva}

\author[1]{Sviatlana A. Ulasevich}

\author[1]{Ekaterina I. Shishatskaya}

\author[1]{Ekaterina V. Skorb}

\author[1]{Pavel S. Zun}

\address[1]{Infochemistry Scientific Center, ITMO University, Lomonosov St. 9, St. Petersburg 191002, Russia}

\address[2]{Institute of Biophysics Russian Academy of Science, Siberian Division, Akademgorodok 50, 660036 Krasnoyarsk, Russia}

\address[3]{National Research Nuclear University MEPhI, Kashirskoe hwy 31, Moscow 115409, Russia}

\begin{abstract}
    Advanced image segmentation and processing tools present an opportunity to study cell processes and their dynamics. However, image analysis is often routine and time-consuming. Nowadays, alternative data-driven approaches using deep learning are potentially offering automatized, accurate, and fast image analysis. In this paper, we extend the applications of \textit{Cellpose}, a state-of-the-art cell segmentation framework, with feature extraction capabilities to assess morphological characteristics. We also introduce a dataset of DAPI and FITC stained cells to which our new method is applied.
\end{abstract}



\begin{keyword}



  Microscopy \sep
  Image analysis  \sep
  Image segmentation \sep
  Bioimaging \sep

\end{keyword}

\end{frontmatter}



\section{Introduction}
The understanding of a successful cell culturing is based on a correct assessment of the samples. Image analysis provides quantitative and qualitative support for assessing cell health; microscopy also identifies nuclear damage, mycoplasma contamination, cell mutations, as well as visual signs of cell differentiation \cite{mey2000development, waisman2019deep, xing2017deep}. However, for the understanding of the biological mechanisms, precise analysis and statistical quantification requires a large amount of image data which constantly increases. This makes manual processing of the image data inefficient and time-consuming. Nowadays, much attention is paid to computerized methods and deep learning technologies as they can significantly improve the efficiency and objectiveness of bioimage processing \cite{wang2019deep, ouyang2018deep}. It should be noted that the microscopy image analysis requires multiple parameters and time-consuming iterative algorithms for processing because they contain a lot of information due to different signal-to-noise ratios \cite{ouyang2018deep}. Advances in cell image analysis present tools to approach different visual characteristics \cite{nketia2017analysis, caicedo2017data, bray2012workflow} with the use of traditional algorithms. 

\begin{figure}[ht]
  \centering
  \includegraphics[scale=0.6]{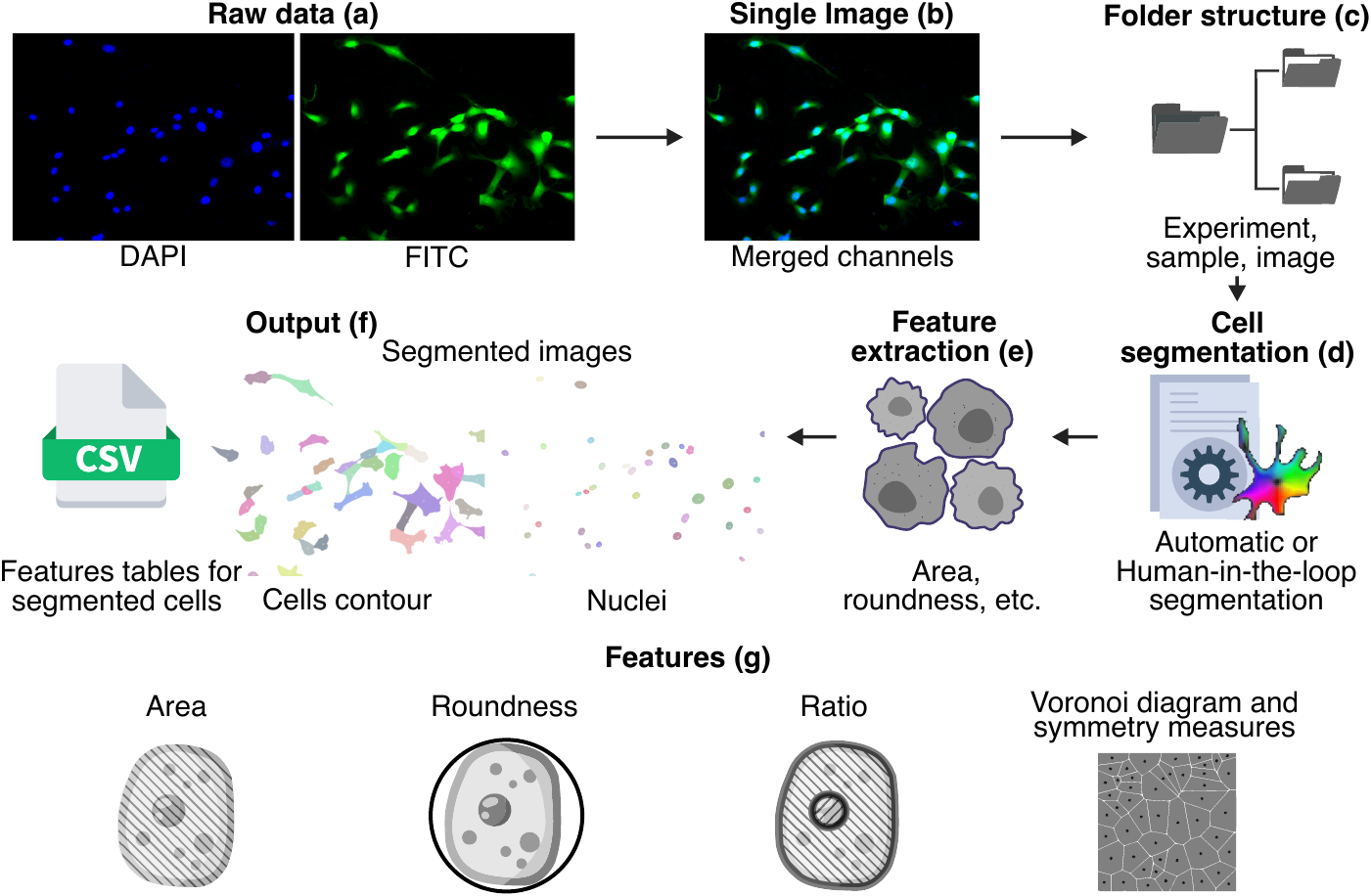}
  \caption{Workflow to obtain metrics from segmented cells. Initial raw data (a) is merged into a single image (b) and organized into sub folders to be processed (c). A cell segmentation procedure is performed using Cellpose (d). We extract metrics (e) according the defined folder structure, and finally we obtain results (f) in the form of images and CSV files containing mentioned metrics (g).}
  \label{figure1}
\end{figure}

A further step is the segmentation and feature extraction of cells from culture images; a process that has been faced with the use of deep learning algorithms \cite{van2016deep, moen2019deep} and image analysis environments \cite{stirling2021cellprofiler, bray2016cell, valente2017simple}. Cell segmentation task is a complex problem that has been approached from simpler solutions, such as image analysis tools and pre-processing techniques like binarization and thresholding, to more recent techniques, e.g. \textit{SuperSegger}\cite{stylianidou2016supersegger}, use both image filtering and neural network resources to correct common image analysis errors. Early machine learning solutions were based on Mask R-CNN architecture\cite{he2017mask}, while recent algorithms like \textit{Stardist}\cite{schmidt2018cell}, \textit{MiSiC}\cite{panigrahi2021misic}, \textit{Cellpose}\cite{pachitariu2022cellpose} and \textit{Omnipose}\cite{cutler2022omnipose} are based on U-net architecture variations. Our approach, the details of which can be seen in Figure \ref{figure1}, begins by selecting an extensible cell segmentation tool with a state-of-the-art performance, to couple it with feature extraction and analysis algorithms. Along with this tool, we also present a novel dataset of DAPI and FITC stained cells.

Detection of nuclei and cells on photos is very important in microscopy image analysis. For example, nucleus or cell recognition can provide significant support for object counting, segmentation, and tracking. Currently, convolutional neural networks (CNNs), fully convolutional networks (FCNs), and stacked autoencoders (SAEs) have all been successfully used for object detection in images, and the locations of objects are often suggested with identified single points near object centroids, which are indicated as seeds or markers \cite{xing2017deep}. Thus, object detection can be formulated as a pixelwise classification problem. Analyzing an input image, the network outputs a probability map, where each pixel value implies the probability of one pixel to be a seed. Therefore, the target objects can be located by seeking local maxima in the generated probability map. In practice, non-maximum suppression is often used to improve the accuracy \cite{xing2017deep, xie2018microscopy, xu2015stacked}

\section{Methods}
\subsection{Polymer production}
Poly-3-hydroxybutyrate-co-hydroxyvalerate (10\% hydroxyvalerate) was synthesized by the batch culture of \textit{Cupriavidus
necator} B-10646, known for high PHA copolymer yields and substrate tolerance, according to the
standard protocol \cite{alexander2008biosynthesis}. Cultivation was carried out in a BioFlo-115 automated laboratory fermentor (Bioengineering AG, Switzerland) under strictly aseptic conditions. Phosphate-buffered
Schlegel medium with glucose as the main carbon source was used. Valeric acid was used as a precursor of valerate. To obtain high polymer yields, a two-stage process was carried out. In the first stage, cells were grown under nitrogen deficiency. In the second stage, cells were cultured in nitrogen-free medium. 

The inoculum was prepared in a shaker incubator InnovaH 44 (New Brunswick
Scientific, U.S.) by resuspending the museum culture \cite{volova2021properties}. Polymer recovery was performed with ethanol for elimination of lipids and fatty acids, followed by polymer extraction in dichloromethane. The dichloromethane extracts were evaporated twice using a rotary evaporator (Rotovapor R-300, Buchi, Germany). Then, the polymer was precipitated with hexane and dried. 

The polymer content was determined using gas chromatography with a chromatograph-mass spectrometer. To purify the polymer, it was dissolved in chloroform multiple times and precipitated using either isopropanol or hexane. Finally, the resulting polymer was dried at 40°C \cite{kiselev2012technical}. 

Analysis of PHA structure, purity and physicochemical properties of P3HBV were determined using a standard set of tests. H 1 NMR spectra of copolymer were recorded in CDCl3 on a Bruker AVANCE III 600 MHz spectrometer (Bruker, Germany) operating at 600.13 MHz. The molecular weight characteristics were determined by a gel permeation chromatography (‘‘Agilent Technologies’’ 1260 Infinity, U.S.) with a refractive index detector and an Agilent PLgel Mixed-C column. Thermal characteristics were fixed by a DSC-1 differential scanning calorimeter (METTLER TOLEDO, Switzerland). XRD structure analysis was carried out employing a D8ADVANCE X-Ray powder diffractometer equipped with a VANTEC fast linear detector, using CuKa radiation (‘‘Bruker, AXS’’, Germany), so as the determination of P3HBV crystallinity \cite{volova2014glucose}. 

P3HBV polymer films of different thicknesses were prepared by casting solution technique. P3HBV was dissolved in chloroform to obtain homogeneous solutions with concentration of 1.0, 1.5, 2.0, 2.5 and 3.0\%. These solutions were gently heated to 35°C for complete dissolution before being cast onto clean, degreased glass surfaces. The films were dried at room temperature for 48 hours in a dust-free environment, allowing the chloroform to evaporate completely and resulting in solid films. Thickness of samples was measured at six points for each film by Legioner EDM-25-0.001 micrometer (Legioner, China) \cite{ghorabe2024insight}. Upper surfaces of the films were used for cell cultivation.

\subsection{Cell culture and image acquisition}
\label{ss:image-aq}

Linear culture of mice fibroblasts NIH 3T3 was used to obtain the visual data of cell adhesion on P3HBV film samples \cite{shishatskaya2022resorbable}. Cells were cultured under sterile conditions in a humidified atmosphere: at 37°C with 5\% $CO_2$ (MCO-19AIC SANYO Electric Co., Ltd., Japan). Dulbecco’s Modified Eagle Medium, DMEM (Gibco, USA) supplemented with 10\% fetal bovine serum, FBS, (HyClone, USA) and antibiotic/antimycotic (Sigma-Aldrich, USA) was used. Cells were detached from the bottom of culture flasks using trypsin (Gibco, USA) when reaching 85–95\% confluence. 

To assess the cytocompatibility of PHA samples, cells were seeded onto sterile polymer films samples at a density of $2 \times 10^4~cells/cm^2$ and cultured for 72 hours. After the incubation, the samples were washed with phosphate-buffered saline and cells were fixed with 4\% paraformaldehyde solution. Cell membranes were permeabilized with 0.2\% Triton-X, and the cytoplasm was stained with fluorescein isothiocyanate, FITC (green) (Sigma-Aldrich, USA), for 1 hour in the dark at room temperature. The nuclei were visualized using 4',6-diamidino-2-phenylindole, DAPI (blue) (Sigma-Aldrich, USA).

Cells were visualized using a Leica DMI8 fluorescent microscope with corresponding LAS X software.  The images were recorded using Leica Microsystem CMS GmbH analyzer (Leica Microsystem, Germany). The scanned area was $1,08~mm^2$ containing $1920 \times 1440$ pixels. The photos were made using N PLAN EPI 10x/0.25 PH1 and N PLAN EPI 20x/0.40 PH1 objectives. Merged cells images are used in the following machine analysis.

\subsection{Cell statistical analysis}

The statistical analysis of cultured cells was performed by the operator using analysis of variance (ANOVA test). As biological objects are variable, the  variance analysis finds dependencies in experimental data by examining the significance of differences in mean values. This method could compare samples based on their means and depict how different these samples are from one another. Besides, the ANOVA test can be used to describe complex relations among variables. The observed variance in a particular variable is partitioned into components attributable to different sources of variation. The simplest variant of the ANOVA test provides a statistical test of whether two or more population means are equal. Thus, ANOVA generalizes the t-test (Student's test) beyond two means.
The cells counted by the human operator were then compared with the cell count obtained by \textit{Cellpose}. The final counting from the software was refined until the standard deviation in the values obtained by the program matched the standard deviation obtained by the human operator.

\section{Results and discussion}

\subsection{Cell segmentation}

We make use of \textit{Cellpose 2.0} \cite{pachitariu2022cellpose} framework to obtain segmented subjects from stained images for cytoplasm (FITC) or nuclei (DAPI). For batch processing of microscopy images, we use \textit{Cellpose} in a noninteractive (command-line) mode. This tool allows the generation of a custom file (.npy) to store partial information as segmentation parameters. We extend the contents of this file to also store segmented masks, in order to allow a manual refinement. 

As a solution for the large number of batches of cell images, we adopted a mixed processing workflow, composed by a script that performs the cell segmentation automatically with pre-defined parameters, and the use of temporary files to manually improve automatic results.

\floatsetup[figure]{style=plain,subcapbesideposition=top}
\begin{figure}
  \sidesubfloat[b]{\includegraphics[width=0.45\linewidth]{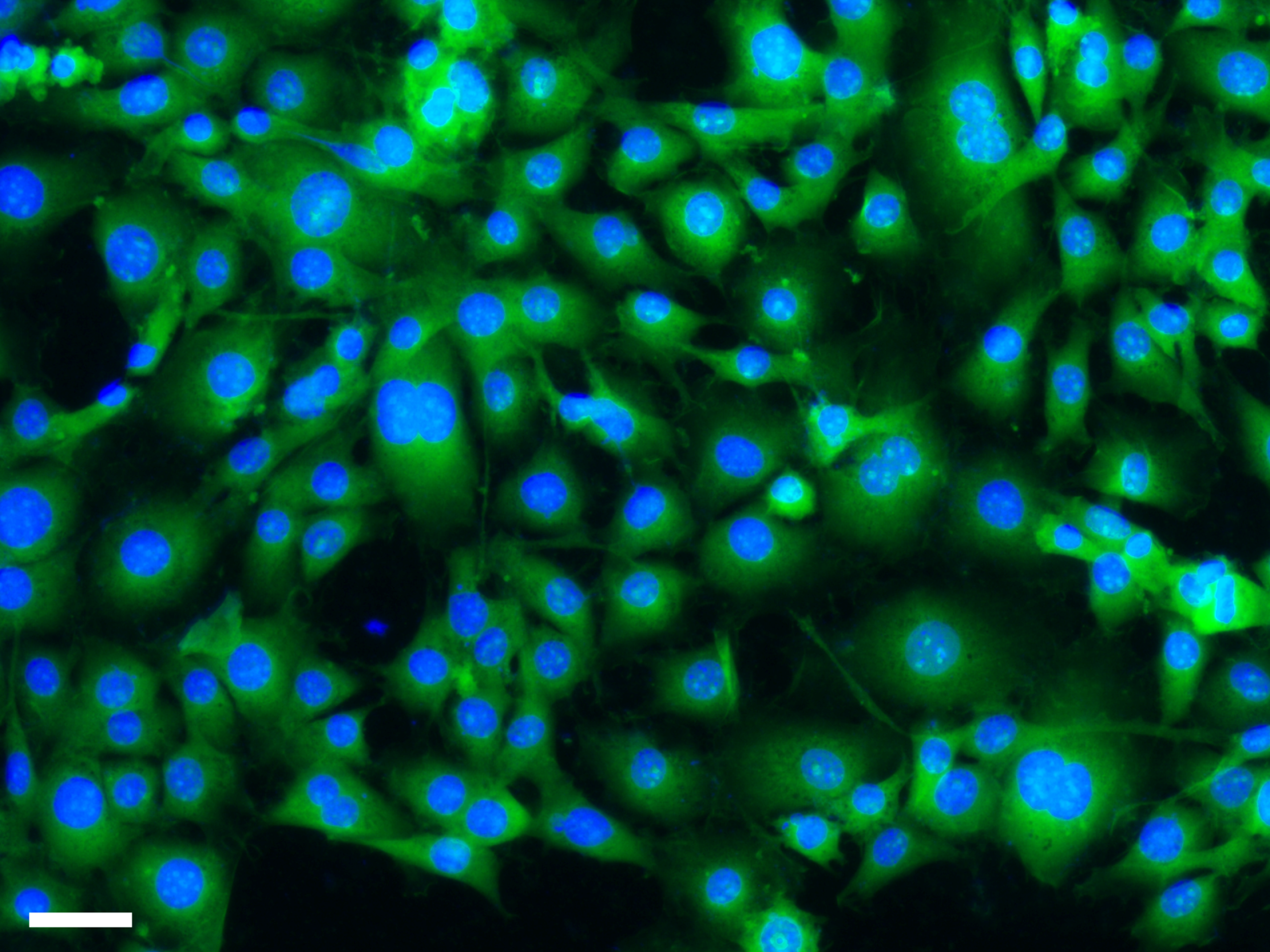}\label{fig:sub1}} \hfil
  \sidesubfloat[b]{\includegraphics[width=0.45\linewidth]{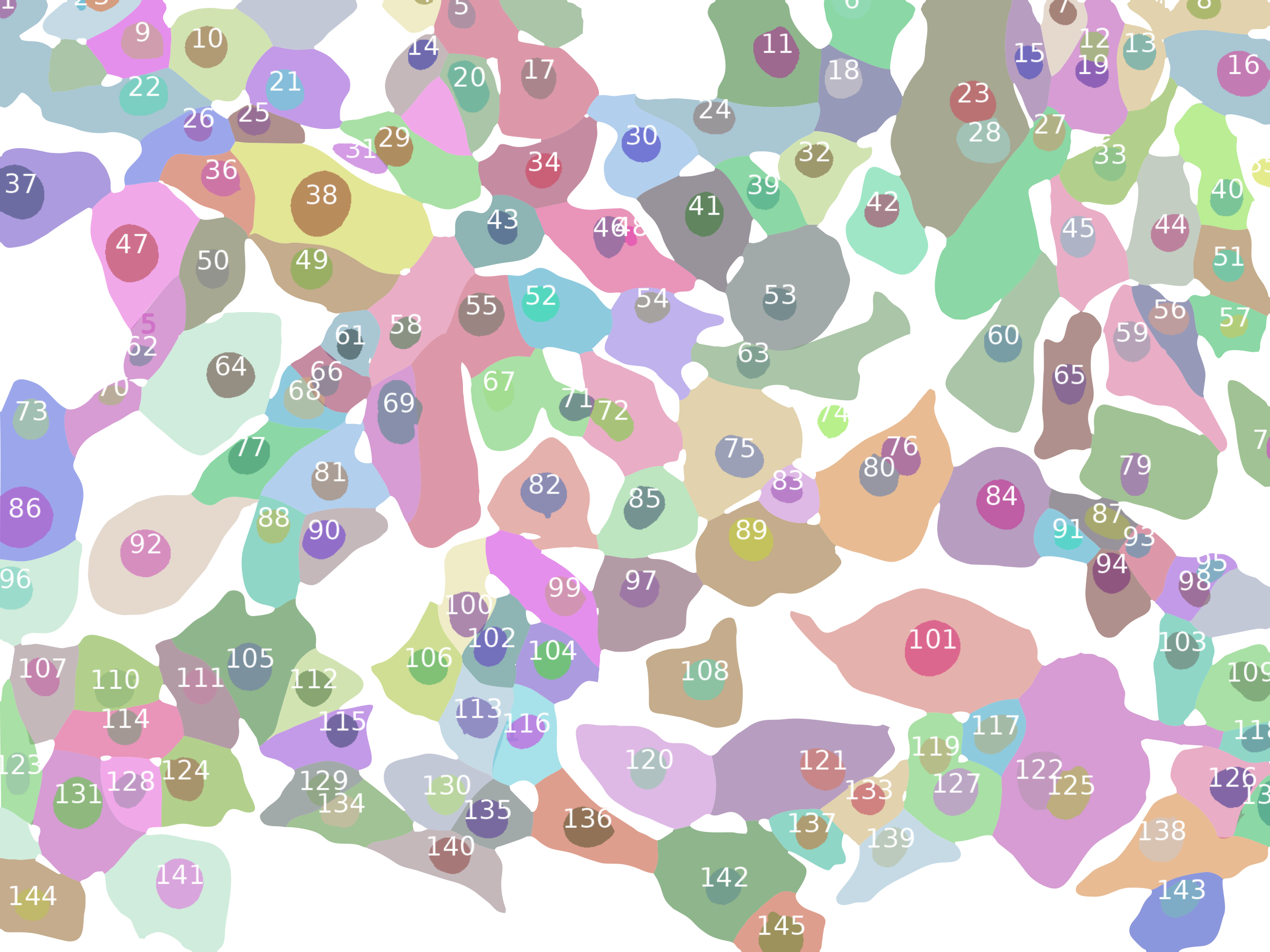}\label{fig:sub2}} \\[\baselineskip]
  \sidesubfloat[b]{\includegraphics[width=0.45\linewidth]{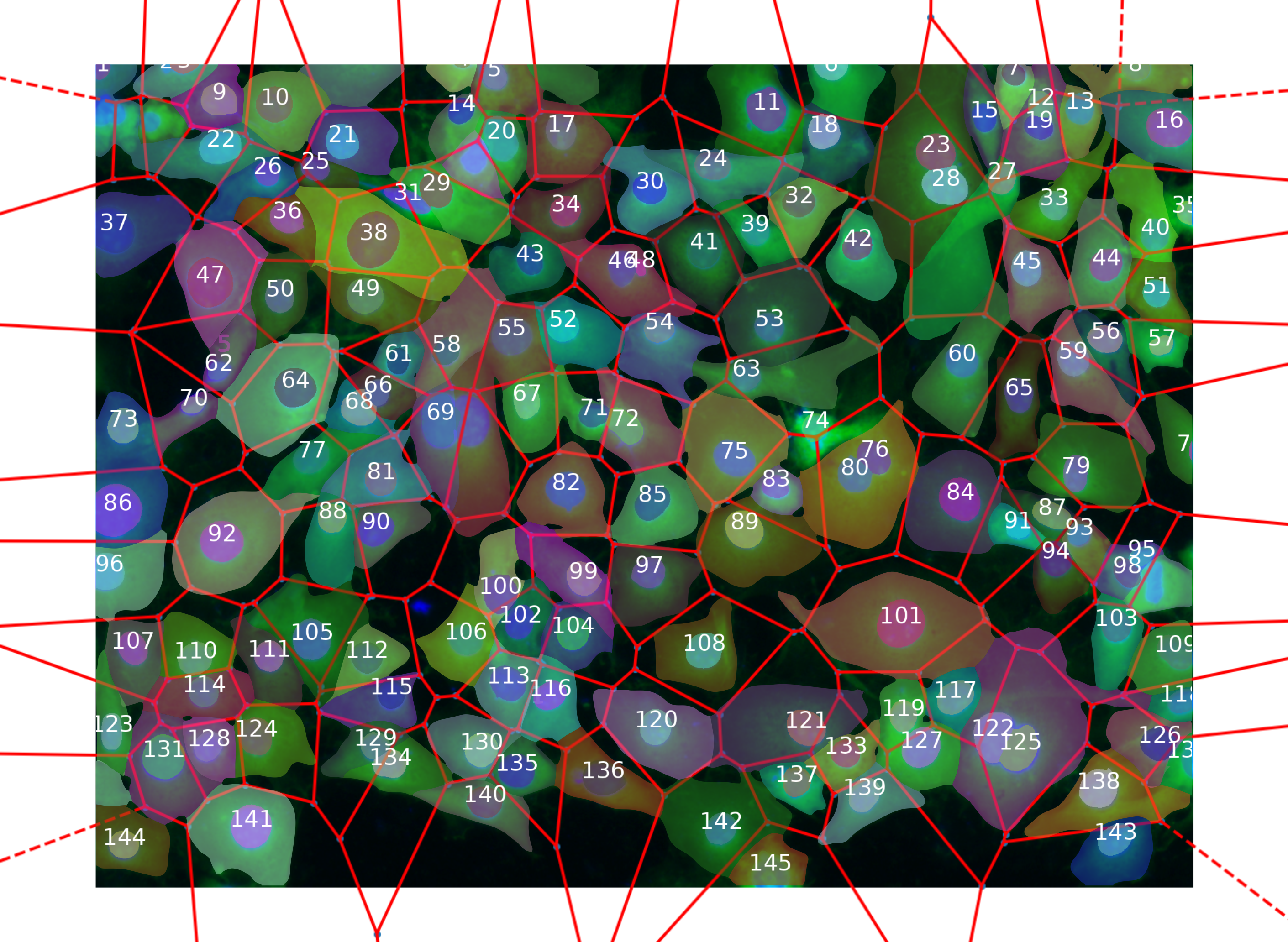}\label{fig:sub3}} \hfil
  \sidesubfloat[b]{\includegraphics[width=0.45\linewidth]{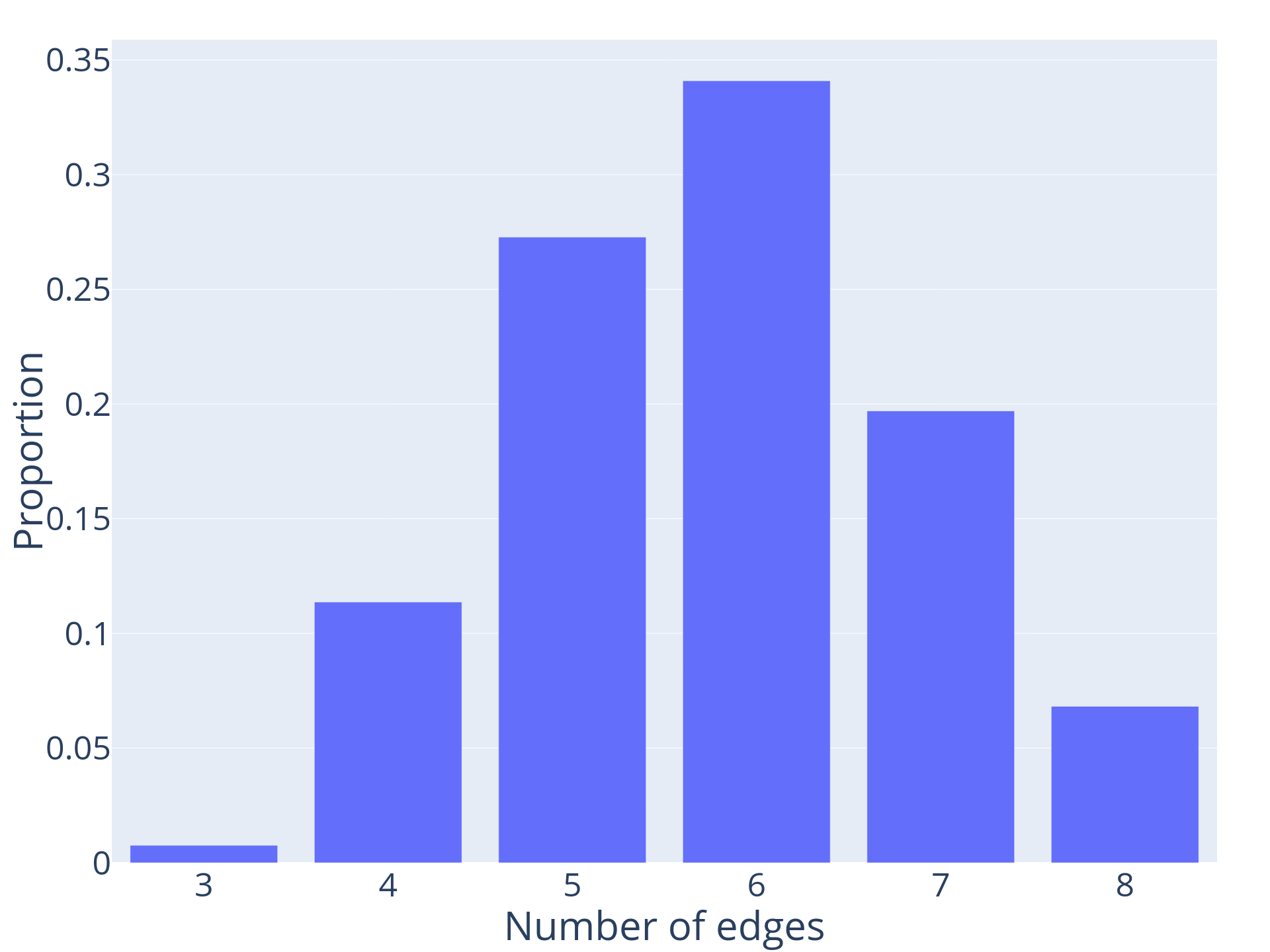}\label{fig:sub4}} \\[\baselineskip]
  \caption{Processing stages of a single image. From the image combined from FITC and DAPI (a), we extract cells contour and nuclei masks (b) for feature extraction. We use center coordinates to plot a Voronoi diagram (c) and get an uniformity reading of it (d). Scale bar, $50~\mu m$.}
  \label{figure2}
\end{figure}

\floatsetup[figure]{style=plain,subcapbesideposition=top}
\begin{figure}
  \sidesubfloat[b]{\includegraphics[width=0.95\linewidth]{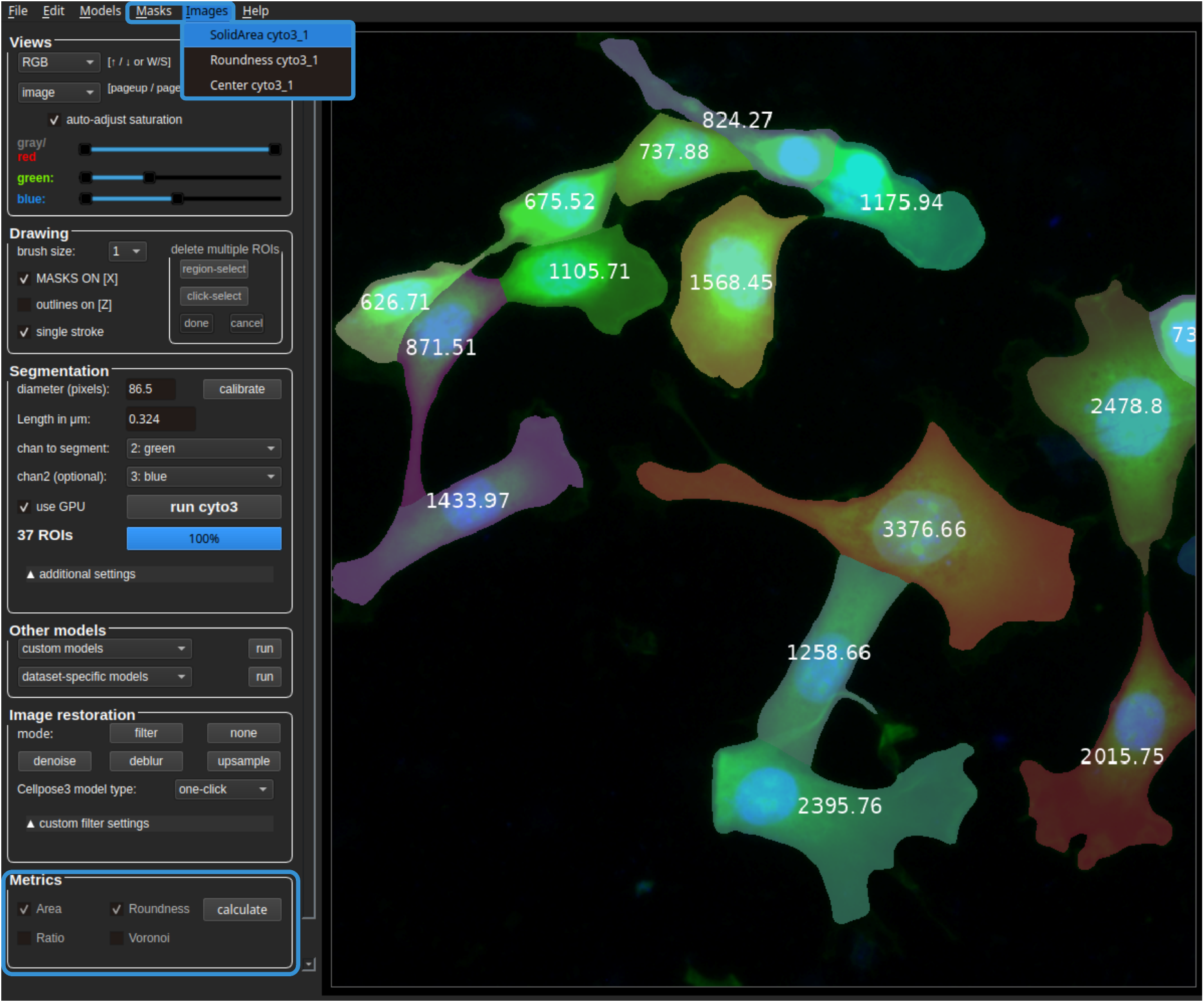}\label{fig:sub1}} \\[\baselineskip]
  \sidesubfloat[b]{\includegraphics[width=0.53\linewidth]{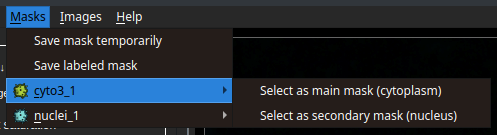}\label{fig:sub3}} \hfil
  \sidesubfloat[b]{\includegraphics[width=0.33\linewidth]{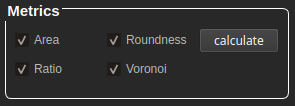}\label{fig:sub4}} \\[\baselineskip]
  \caption{Updated \textit{Cellpose} GUI. A new feature extraction tools marked in cyan, showing segmented cells with their respective area in $\mu m^2$ (a). Two new tabs, Masks and Images, are added to save segmented subjects and to visualize features respectively (b). We added the morphological features we can select to extract from the segmented subjects.}
  \label{figure3}
\end{figure}

\subsection{Implementation of features extraction}

After a segmentation is performed, we proceed to extract characteristics from the segmented cells, which can be used to assess the biocompatibility of the substrate and the viability and state of the cells (see Figure \ref{figure2}). For this study, we make use of \textit{DIP} and \textit{scipy} libraries, which includes algorithms for image analysis. We select relevant features that can be used for further interpretation, which are listed below. When we refer to a subject detected in the image, this indicates either an area of cytoplasm belonging to a single cell or a nucleus.

\begin{itemize}
  \item Total amount of subjects in an image, which is the count of segmented cells per channel.
  \item Area of a subject ($\mu m^2$) is calculated using its chain code (contour). Each subject must be a  connected component with a unique label. The area is calculated as pixels $(px)$ primarily, then converted to square micrometers $(\mu m^2)$ using the dimensions of the image.
  \item Roundness ($R$) of a subject (0.0 - 1.0), having 1.0 for a perfect circle. It is calculated as 
  \begin{equation*}
      R = \frac{4\pi a}{p^2}
  \end{equation*}
  
  where $a$ is the area and $p$ is the perimeter.
  \item The ratio between the sizes of the cell (stained cytoplasm area $C_i$) and its nucleus ($N_i$), calculated as
  \begin{equation*}
     \frac{C_i}{N_i} 
  \end{equation*}
  
  per each pair $i$ of cell and nucleus.
  \item Percentage of the area of the image covered by cells or nuclei, calculated as 
  \begin{equation*}
     \sum\limits_{i=0}^n A_i / A_{T} 
  \end{equation*}
  
  where $A_i$ is the area of a cell or a nucleus and $A_T$ is the total area of the image.
  \item Coordinates of the center of each subject in an image, as a relative position from the origin on the top left corner.
  \item Voronoi diagrams based on the center coordinates of nuclei, calculated and plotted with the \textit{Voronoi} method from \textit{scipy.spatial}.
  \item Voronoi entropy ($S_{Vor}$) from the Voronoi diagram, calculated as 
  \begin{equation*}
     S_{Vor}=-\sum\limits_{i=0}^{n} (p_i \log p_i )
  \end{equation*} 
  
  where $p_i$ is the proportion of each class.
  \item Continuous Symmetry Measure (CSM) from the Voronoi diagram, we obtain a normalized value calculated as 
  \begin{equation*}
     CSM=\sum\limits_{i=0}^{n} \frac{|M_i - \hat{M}_i|^2}{nS_i}
  \end{equation*}

  where $M_i$ is the point of a polygon, $\hat{M}_i$ is the corresponding point of a symmetrical polygon, $n$ is the number of vertices in a polygon and $S_i$ is the area of a symmetrical polygon.
\end{itemize}

\begin{figure}[ht]
  \centering
  \includegraphics[scale=0.30]{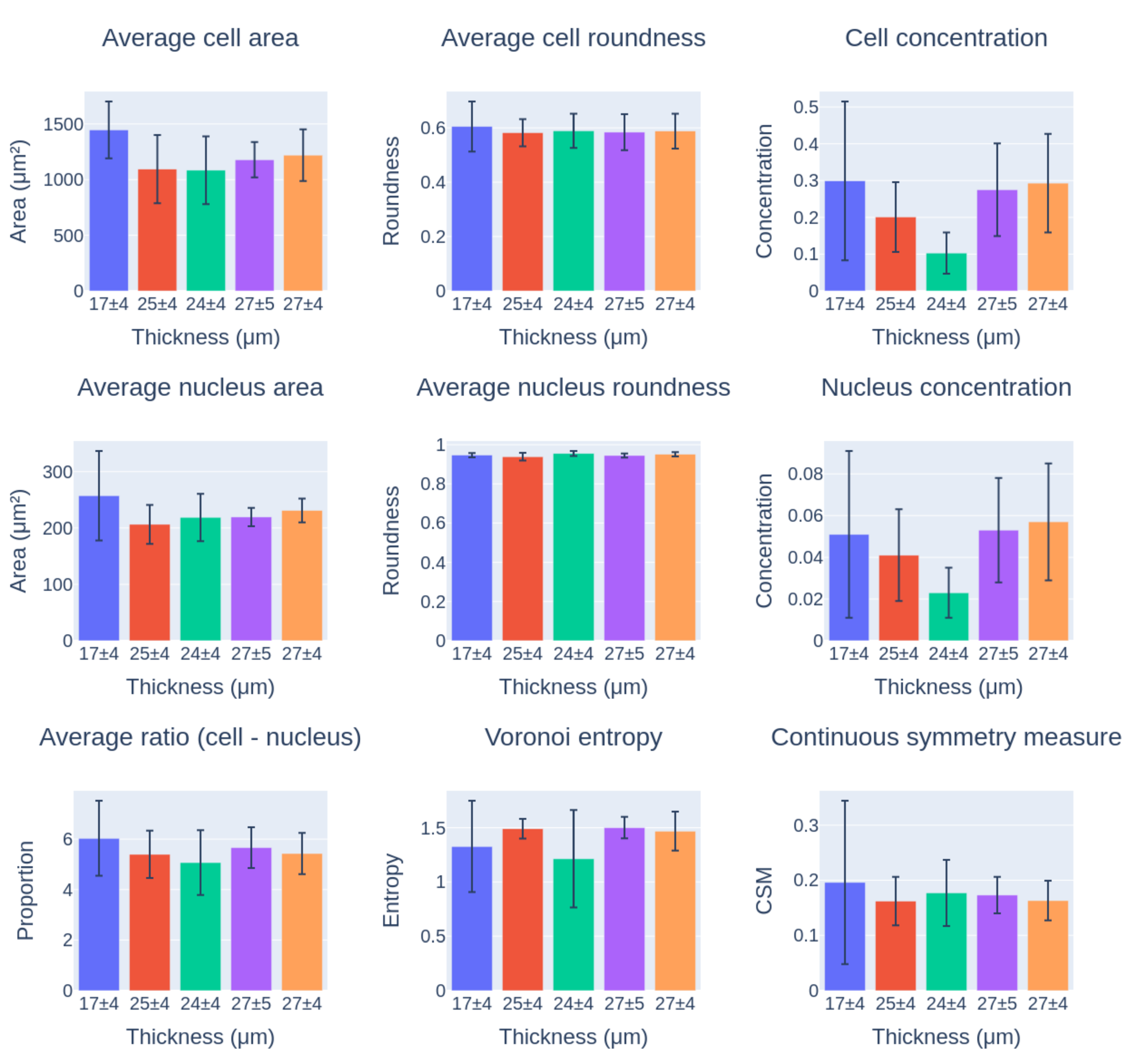}
  \caption{Features extracted from cells and nuclei according to their thickness level.}
  \label{fig:figure4}
\end{figure}

\begin{table}[!ht]
    \centering
    \caption{Values of features extracted per subject (cell, nucleus and miscellaneous) from \cref{fig:figure4}}
    \label{tab:results}
    \begin{tabular}{|c|c|c|c|c|c|c|c|c|}
    \hline
        \multicolumn{3}{|c|}{Cytoplasm} & \multicolumn{3}{c}{Nucleus} & \multicolumn{3}{|c|}{Misc.} \\ \hline
        \makecell{Area} & \makecell{Roundness} & \makecell{Conc.} & \makecell{Area} & \makecell{Roundness} & \makecell{Conc.} & \makecell{Ratio} & Entropy & CSM \\ \hline
        1446.343 & 0.605 & 0.299 & 257.283 & 0.947 & 0.051 & 6.03 & 1.327 & 0.196 \\ \hline
        1093.387 & 0.582 & 0.201 & 206.255 & 0.939 & 0.041 & 5.395 & 1.49 & 0.162 \\ \hline
        1083.944 & 0.589 & 0.103 & 218.712 & 0.955 & 0.023 & 5.069 & 1.213 & 0.177 \\ \hline
        1178.092 & 0.584 & 0.275 & 219.279 & 0.945 & 0.053 & 5.663 & 1.5 & 0.173 \\ \hline
        1219.777 & 0.588 & 0.293 & 230.878 & 0.951 & 0.057 & 5.429 & 1.468 & 0.163 \\ \hline
    \end{tabular}
\end{table}


The tool we developed was used to analyze the dataset outlined in \cref{ss:image-aq}. The features were obtained for five representative images of each of the three films of each of the five different thicknesses. The average values of the measured features for each thickness, are reported in \cref{fig:figure4} and \cref{tab:results}. The detailed results for each film as well as the raw and segmented images can be found in the supplementary materials. The project can be accessed at the public repository \url{https://github.com/ITMO-MMRM-lab/cellpose}, where usage and installation instructions are available. We also provide a dataset composed by raw images used for our experiments, along with their masks \url{https://zenodo.org/doi/10.5281/zenodo.13378558}.

\subsection{Discussion}

To detect a cell or nucleus, it is important to pay attention not only to its shape, but also to its size. In this regard, \textit{Cellpose} identifies objects and compares them on average by area, discarding values that deviate from the target value (automatically calculated or manually set). In the present work, a human operator also adjusted these values.

However, during mitosis, the size of the cell and its nucleus can increase significantly \cite{weaver2005decoding}, which can lead to detection errors. In this regard, we also decided to pay attention to the ratio of the area of the nucleus and cytoplasm (see \cref{fig:figure4}). In addition, we are planning to use \textit{Cellpose+} for mitosis detection in the future. For these purposes, we will additionally use a dataset of NIH 3T3 cells stained by hematoxylin and eosin. It has already been shown \cite{prezja2022h, couture2022deep, lu2016multi} that this type of cell staining separates fragments of the nucleus by color and can be used in deep learning methods based on pixel analysis and imaging such as \textit{Cellpose}.
It should be noted that the counting of nuclei can also be used to estimate the cell density on samples (\cref{fig:figure4}). Assuming that the number of cells on the sample corresponds to the number of nuclei and knowing the area of the analyzed image, it is possible to calculate the cell density (in cells per unit area), which characterizes the number of cells on the sample and can indirectly serve to assess the biocompatibility of the samples. By combining statistical analysis methods (for example, ANOVA) in the resulting method, we can estimate the cell density on samples quite accurately. 
It is known that cells are unevenly distributed on samples with different surface rigidity, which often correlates with the coating thickness and the nature of the material \cite{dalby2014harnessing, donnelly2023cell}. Having tested \textit{Cellpose+} on images of cells on the surface of polyhydroxyalkanate films with different thickness and elasticity, we observed a correlation between the sample thickness and the cell density on the sample. 

The data obtained by \textit{Cellpose+} were compared with experimental data processed by a human operator. It should be noted that the presented method allows you to quickly and efficiently determine the cell density per square millimeter or centimeter of the sample.
It should also be noted that this method can be easily applied to assess cell viability. The most well-known is Live/Dead staining using acridine orange and propidium iodide \cite{boulos1999live, mahaling2022azithromycin}. Cell viability in samples is defined as the ratio of the number of nuclei stained green (indicating living cells) to the number of nuclei stained red, corresponding to the total number of cells in the sample.
Thus, \textit{Cellpose+} can be easily adapted for this technique if you replace the recognition of nuclei in the blue channel (DAPI) with the orange color (RHOD) channel.

It should be mentioned that color and texture-based methods have some limitations as they need steady color/texture appearance for the individual nuclei to work optimally. For example, the initial version of \textit{Cellpose} worked fine on well stained images, whereas due to the variations of staining and cell preparation methods, a number of images that were stained poorly, resulted in detection’s fail on unfairly and fuzzily recorded images. 
We have taken into account some noisy edge pixels outside the true nuclei on the edge segments because of the noise and extraneous tissue components in the image. Removing these signals is very important as the estimated number of nuclei may fall outside the true ones, which leads to an erroneous determination of which pixels are either valid or invalid for analysis. In our experiments, the number of such cases was small.

It should be noted that cell shape analysis is very important, since it allows us to suggest future differentiation of the cell. According to literature review \cite{archer1982cell, kilian2010geometric, hoffman2024image, madrigal2023epigenetic, kopf2016ultrasonically}, some cell lines are known to have a certain morphology before cell differentiation. 
For instance, control MSCs cells without pharmacological agents show 72\% of MSC cells on the flower shape have an adipogenic fate compared to 67\% of cells on the star shape showing an osteogenic fate \cite{kilian2010geometric}. In particular, star-like cells are found to promote expression of noncanonical Wnt/Fzd signaling molecules (including down-stream effectors RhoA and ROCK previously shown to be involved in osteogenesis \cite{kilian2010geometric, hoffman2024image}. There is the same tendency for C2C12 cells which are capable to differentiate along several lineages and is used as a model system for osteogenic and myogenic differentiation \cite{kopf2016ultrasonically}.
The cells that adopted a star-like morphology tended to osteogenic differentiation while the cells having an elongated spindle-like cytoskeletal structure moderately exhibited myogenic differentiation \cite{kopf2016ultrasonically}. The geometric shape of a cell may be assigned to the mechanochemical signals and paracrine/autocrine factors that can direct MSCs to appropriate fates \cite{hoffman2024image}.

Another point why it is important to assess the shape of a cell, can be to predict its migration along the sample. It is known that when a cell moves, its morphology changes, becoming more elongated, rounded or spread-out. In addition to the cell skeleton, the shape of the cell nucleus also changes. Thus, in this work, we lay a foundation for future studies of complex cell analysis.

\section{Conclusions}
Given the relevance of segmentation procedures for cell imaging using deep learning models,
the integration of morphological analysis algorithms to a high-end cell segmentation framework
became a necessity that we satisfied. The presence of a solid GUI makes the framework accessible
for users with different backgrounds. Having this as priority, we selected the Cellpose project
to extend it with feature extraction algorithms. The repository of our Cellpose version was
forked directly from the main source, so it is continuously synchronized. Although Cellpose
allows 3D segmentation, our algorithms cover only 2D image analysis. Regarding the dataset,
we provide the analyzed images samples with all their available features extracted, as well as
the rest of the raw samples with their masks. For future steps, we would like to extend the number
of features extracted, as well as improve the GUI according to the increase of functionalities, and
then, provide the possibility to create and use plug-ins according to specific necessities.

\section{Acknowledgements}

The research was carried out within the project No. 24-075-61691-1-4119-000082. The ITMO Fellowship and Professorship Program is acknowledged for infrastructural support.





\bibliographystyle{vancouver}
\bibliography{manuscript}

\end{document}